\documentclass[]{article}
\usepackage{amsmath}
\usepackage{amsthm}
\usepackage{amssymb}
\usepackage{authblk}
\usepackage{mathtools}
\usepackage{makecell}
\usepackage{physics}
\usepackage{hyperref}
\usepackage{bm}\newcommand{\ind}{\bm{1}}
\usepackage{multirow}
\usepackage[table]{xcolor}
\usepackage{booktabs}
\usepackage[utf8]{inputenc}
\usepackage{adjustbox}
\usepackage{caption}
\usepackage{rotating}
\usepackage{graphicx}
\graphicspath{{fig/}}
\usepackage[margin=3cm]{geometry}

\DeclarePairedDelimiter{\floor}{\lfloor}{\rfloor}
\DeclarePairedDelimiter{\ceil}{\lceil}{\rceil}
\DeclareUnicodeCharacter{2713}{\checkmark}
\makeatletter
\newcommand{\printfnsymbol}[1]{%
	\textsuperscript{\@fnsymbol{#1}}%
}
\makeatother

\providecommand{\keywords}[1]
{
	\small	
	\textbf{\textit{Keywords:}} #1
}

\title{A flexible integer linear programming formulation for scheduling
	clinician on-call service in hospitals}

\author[a, b]{David Landsman}
\author[a]{Huiting Ma\thanks{Contributed equally to this work.}}
\author[a, c]{Jesse Knight\printfnsymbol{1}}
\author[d]{Kevin Gough}
\author[a, c, d, e]{Sharmistha Mishra\thanks{Corresponding author (\texttt{sharmistha.mishra@utoronto.ca})}}

\date{}

\affil[a]{MAP Centre for Urban Health Solutions, St.\ Michael's Hospital, 
	Unity Health Toronto, Toronto, ON, Canada}
\affil[b]{Department of Computer Science, University of Toronto, Toronto, ON,
	Canada}
\affil[c]{Institute of Medical Sciences, University of Toronto, Toronto, ON,
	Canada}
\affil[d]{Department of Medicine, Division of Infectious Disease, St.\ Michael's
	Hospital, Unity Health Toronto, Toronto, ON, Canada}
\affil[e]{Institute of Health Policy, Management and Evaluation, Dalla Lana
	School of Public Health, University of Toronto, Toronto, ON, Canada}

\begin{document}
	\maketitle
	
	\begin{abstract}
		Scheduling of personnel in a hospital environment is vital to improving the
service provided to patients and balancing the workload assigned to clinicians.
Many approaches have been tried and successfully applied to generate efficient
schedules in such settings. However, due to the computational complexity of the
scheduling problem in general, most approaches resort to heuristics to find a
non-optimal solution in a reasonable amount of time. We designed an integer
linear programming formulation to find an optimal schedule in a clinical
division of a hospital. Our formulation mitigates issues related to
computational complexity by minimizing the set of constraints, yet retains
sufficient flexibility so that it can be adapted to a variety of clinical
divisions. \\

We then conducted a case study for our approach using data from the Infectious
Diseases division at St. Michael's Hospital in Toronto, Canada. We analyzed and
compared the results of our approach to manually-created schedules at the
hospital, and found improved adherence to departmental constraints and clinician
preferences. We used simulated data to examine the sensitivity of the runtime of
our linear program for various parameters and observed reassuring results,
signifying the practicality and generalizability of our approach in different
real-world scenarios.

	\end{abstract}
	
	\section{Introduction}\label{sec:introduction}
	Hospital departments must allocate and use their limited resources efficiently
in order to provide a high quality of care for their patients.
In particular, on-call schedules for
a fixed number of health-care providers are central to the efficient running of
hospitals. Carefully allocated schedules should balance
sufficient staff with workload
to maximize quality of care. It is common for on-call schedules in
hospitals to be created manually. However,
manually-created schedules are subject to three problems.
First, when there are a large number of clinicians,
or the constraints that need to be satisfied
by the schedule are complex, it becomes infeasible
to find a satisfying schedule by hand. 
Second, when creating a schedule manually it is difficult to ensure
that all constraints are met while also trying to satisfy all staff
preferences.
Third, manual scheduling is often time-consuming even for relatively small 
departments, and can take up time and resources that are better used
for improving patient care.
For these reasons, it is important to develop automated methods that
can efficiently generate schedules that satisfy the given constraints.

Automated methods to generate schedules have been studied and applied in many
industries, including
transportation~\cite{aickelin_improved_2006, goel_truck_2012, gunther_combined_2010},
manufacturing~\cite{al-yakoob_mixed-integer_2007, al-yakoob_column_2008, alfares_simulation_2007},
retail~\cite{chapados_retail_2011, nissen_automatic_2010}, and
military~\cite{horn_scheduling_2007, laguna_modeling_2005}.
Of special interest to a clinician scheduling
problem are the approaches to scheduling nurses, who often work in shifts. In the
nurse scheduling problem, the goal is to find an optimal assignment of nurses to
shifts that satisfies all hard constraints (such as hospital regulations),
and as many soft constraints (such as nurse preferences) as possible.
Hard constraints must be satisfied by any candidate solution to the nurse scheduling problem,
while soft constraints can be used to rank the candidate solutions. 
For instance, a nurse scheduling problem may include a hard constraint to assign
at most a single shift for each nurse per day. It can also incorporate 
nurse preference for shift time (that is, day versus night shifts) as a soft constraint
that is meant to optimize the schedule, but is not guaranteed to be fulfilled.
A wide variety of approaches, including exact and heuristic approaches, have been
used to solve the nurse scheduling problem:
integer linear programming~\cite{azaiez_0-1_2005, trilling_nurse_2006, widyastiti_nurses_2016},
network flows~\cite{el_adoly_new_2018},
genetic algorithms~\cite{aickelin_exploiting_2000, jan_evolutionary_2000, kawanaka_genetic_2001},
simulated annealing~\cite{jaszkiewicz_metaheuristic_1997}, and
artificial intelligence~\cite{abdennadher_nurse_1999, li_hybrid_2003}.
A comprehensive
literature review of these and other methods applied to nurse scheduling is
presented in~\cite{burke_state_2004}.

Many of the approaches to nurse scheduling were designed to satisfy the
requirements of a specific hospital department, which results in a large number of
variables and constraints to be incorporated into the problem formulation. While
these department-specific approaches allow end-users to find precise schedules
that satisfy the needs of that department and the preferences of nurses and
clinicians in that department, they are difficult to adapt to other
departments.
Moreover, the large number of variables and constraints also leads to
computational complexity issues~\cite{goos_complexity_1996}, especially when
trying to find the most optimal solution. In particular, difficult instances of these formulations
become impossible to solve in a reasonable amount of time. 
In this paper, we tackle the clinician 
scheduling problem arising from a case study of one clinical
division, providing two different services (general infectious
disease (ID) consults; and HIV consults) at St.\ Michael's Hospital in
Toronto, Canada. The clinician scheduling problem involves creating
a yearly schedule that assigns clinicians to on-call work on a weekly basis,
while ensuring a fair and balanced workload.
Our goals in this paper are to (1) present an integer linear programming 
(ILP) formulation for our problem, and
describe the flexibility of this formulation for solving similar problems;
(2) compare the performance of our tool for solving the ILP formulation to the
results of a manual approach; 
and (3) analyze the robustness of this approach
in difficult instances of the problem, by exploring the change in runtime with
changes to:
the number of clinicians,
the number of services provided by the department,
the number of requests per clinician, and
the time-horizon of the schedule.

We begin by describing the details of the problem in Section~\ref{sec:problem}, and presenting
our ILP formulation in Section~\ref{sec:methods}. Next, we compare the results
of our formulation to manually-created schedules, and evaluate the performance
of the algorithm on simulated data in Section~\ref{sec:experiments}. Finally, we
discuss and interpret the results in Section~\ref{sec:discussion}.
	\section{Problem}\label{sec:problem}
	At St.\ Michael's Hospital, the division of infectious diseases offers
separate but concurrent services for general ID consultations and for HIV consultations. Each
service provides clinical care throughout the year, during regular working hours
and on weekends and holidays. The schedule is created in advance,
outlining all work-week and weekend shifts for the full year.
In the yearly schedule clinicians are assigned to ``blocks'' of
two consecutive regular work weeks and individual weekends. Apart
from long (holiday) weekends, a work week starts on Monday at 8 A.M. and ends on
Friday at 5 P.M. Accordingly, weekend service starts on Friday at 5 P.M., and
ends on Monday at 8 A.M. During the weekend, ID and HIV consultation services
are combined and provided by one clinician. During the regular work week the ID
and HIV services are led by one clinician each.
Therefore, our objective is to assign a single clinician to cover
each service for each block and each weekend of the year, while 
additionally ensuring a balanced workload.

In most scheduling problems, the constraints can be divided into hard and soft
constraints. Hard constraints must be satisfied by any candidate solution, while
soft constraints can be used to select a more favourable solution from a set of
candidate solutions. Typically, soft constraints are therefore encoded as objective
functions which are
maximized when finding a solution. In the case of the clinician
scheduling problem, we chose departmental regulations and workload balance as hard constraints,
while clinician preference and requests serve as soft constraints.
After the schedule is generated, clinicians may
exchange certain weeks or days throughout the year, to fulfill any missed requests.

The following are the departmental and workload constraints placed on the clinician assignments. 
First, each clinician
must work between a minimum and maximum number of blocks of each service during the year.
For instance, one clinician might have to
provide 3-5 blocks of general ID service and 2-3 blocks of HIV service
throughout the year. These limits may be different for each clinician,
and they may change from year to year as
the number of clinicians in the department changes. Second, the schedule
should not assign a clinician to work for two consecutive blocks or two
consecutive weekends. The schedule should also distribute regular
weekends and holiday weekends each equally among all clinicians.

In addition to balancing the workload among clinicians, the schedule
should accommodate their preferences. Clinicians provide their requests for
time off before schedule generation so that the requests may be integrated
into the schedule. Clinicians may specify days, weeks or weekends off, with the
understanding that any blocks overlapping with their request will be assigned to
a different clinician where possible. For example, if a clinician only requests
a given Monday and Tuesday off, the schedule should avoid assigning the
entire block to that clinician. Clinicians also typically prefer to have their weekend and
block assignments side by side, so the schedule should accommodate this where possible.
A summary of the outlined constraints is given in
Table~\ref{tbl:constraint-summary}.

\begin{table}[h]
	\centering
	\caption{Summary of the constraints for the clinician scheduling problem}%
  \label{tbl:constraint-summary}
	\begin{tabular}{ l c l l }
		\toprule
		\textbf{Constraint Name} & \textbf{Abbreviation} & \textbf{Description}                                                                                                                 & \textbf{Type} \\ \midrule
		Block Coverage                                                                  & BC                    & \makecell[l]{each service needs to have
			exactly \\ one clinician that covers any given block}                                     & Hard          \\ \hline
		Weekend Coverage                                                                & WC                    & \makecell[l]{every weekend needs to have
			exactly \\ one clinician that covers it}                                                 & Hard          \\ \hline
		Min/Max                                                                         & MM                    & \makecell[l]{for a given service, each
			clinician can only \\ work between the minimum and maximum \\ number of allowed
			blocks} & Hard          \\ \hline
		No Consecutive Blocks                                                           & NCB                   & \makecell[l]{any clinician should not work \\
			two consecutive blocks, across all services}                                        & Hard          \\ \hline
		No Consecutive Weekends                                                         & NCW                   & \makecell[l]{any clinician should not work two
			consecutive \\ weekends}                                                           & Hard          \\ \hline
		Equal Weekends                                                                  & EW                    & \makecell[l]{weekends should be equally
			distributed \\ between clinicians}                                                        & Hard          \\ \hline
		Equal Holidays                                                                  & EH                    & \makecell[l]{long weekends should be equally
			distributed \\ between clinicians}                                                   & Hard          \\ \hline
		Block Requests                                                                  & BR                    & \makecell[l]{each clinician can request to be
			off service \\ during certain blocks throughout the year}                           & Soft          \\ \hline
		Weekend Requests                                                                & WR                    & \makecell[l]{each clinician can request to be
			off service \\ during certain weekends throughout the year}                         & Soft          \\ \hline
		Block-Weekend Adjacency                                                         & BWA                   & \makecell[l]{the block and weekend
			assignments of a given \\ clinician should be adjacent}                                        & Soft          \\ \bottomrule
	\end{tabular}
\footnotesize\raggedright
Hard constraints must be satisfied by any candidate schedule. 
Soft constraints are optionally satisfied, and are used to rank the set of candidate solutions.
\end{table}
	\section{Methods}\label{sec:methods}
	In this section, we present an application of integer linear programming to solve the clinician
scheduling problem presented in Section~\ref{sec:problem}. 
An integer linear program (ILP) consists of a linear objective function and linear constraints,
with integer-valued variables. The optimal solution to the ILP must lie within the space
defined by the constraints while also maximizing the objective value.
First, we describe
the sets, indices and variables necessary for the formulation of the problem. We
then write the constraints given in Table~\ref{tbl:constraint-summary} as
linear functions of the variables, and define the objective function of the ILP.

\subsection{Sets and Indices}\label{sec:meth-sets-indices}
We denote the set of all services
that clinicians in a single department can provide as $\mathcal{S}$. 
The set of all clinicians in the department is denoted as
$\mathcal{C}$. The sets of blocks and weekends
are denoted as $\mathcal{B}$ and $\mathcal{W}$ respectively. The block size
used in our experiments is 2 weeks, but the following LP formulation does not
require a particular size for blocks, and so it can be adapted for other cases.
A subset of weekends are denoted $\mathcal{L} \subset \mathcal{W}$, corresponding to statutory
long/holiday weekends such as the Thanksgiving weekend in the United States and Canada. Lastly,
the block and weekend time-off requests of clinicians are denoted as the following
subsets of all blocks and weekends: $\mathcal{U}_c \subset \mathcal{B}$ and $\mathcal{V}_c \subset \mathcal{W}$, respectively.
For instance, if clinician $c$'s time-off requests intersect with blocks 1 and 2, and weekend
1, then $\mathcal{U}_c = \{1, 2\}$ and $\mathcal{V}_c = \{1\}$. Table~\ref{tbl:sets-indices} presents a summary of the sets and indices described. 

\begin{table}[h]
	\centering
  \caption{Description of sets and indices in the problem}%
  \label{tbl:sets-indices}
	\begin{tabular}{ c c l }
		\toprule
		\textbf{Set}                         & \textbf{Index} & \textbf{Description}  
		\\ \midrule
		$\mathcal{S} = \{1, \ldots, S \}$    & $s$            & services              
		\\
		$\mathcal{C} = \{1, \ldots, C \}$    & $c$            & clinicians            
		\\
		$\mathcal{B} = \{1, \ldots, B \}$    & $b$            & blocks                
		\\
		$\mathcal{W} = \{1, \ldots, W \}$    & $w$            & weekends              
		\\
		$\mathcal{L} \subset \mathcal{W}$    &                & long weekends         
		\\
		$\mathcal{U}_c \subset \mathcal{B}$ &                & block requests of
		clinician $c$   \\
		$\mathcal{V}_c \subset \mathcal{W}$ &                & weekend requests of
		clinician $c$ \\
    \bottomrule
	\end{tabular}
	
\end{table}

\subsection{Variables}\label{sec:meth-variables}
Since each clinician may be assigned to work on any service, during any block of
the year, we denote such an assignment as a binary variable $X_{c, b, s} \in \{0,1\}$.
A value of 1 indicates that clinician $c$ is assigned to service $s$ during block $b$,
while a value of 0 indicates they are not assigned.
Weekend assignments are similarly defined using a binary
variable $Y_{c, w} \in \{0,1\}$, but without a service index, as clinicians are expected to
provide all services during the weekends. We then define
$m_{c, s}$ and $M_{c, s}$ to represent the minimal and maximal number of blocks of service $s$
that clinician $c$ is required to work during the year.
Table~\ref{tbl:variables-constants} presents a summary of the constants and variables
in the problem.

\begin{table}[h]
	\centering
  \caption{Description of variables and constants in the problem}%
  \label{tbl:variables-constants}
	\begin{tabular}{ c l }
		\toprule
		\textbf{Name}              & \textbf{Description}                             
		\\ \midrule
		$X_{c, b, s} \in \{0, 1\}$ & assignment of clinician $c$ to service $s$ on
		block $b$            \\
		$Y_{c, w} \in \{0, 1\}$    & assignment of clinician $c$ on weekend $w$       
		\\
		$m_{c, s}$                 & minimum number of blocks clinician $c$ should
		cover on service $s$ \\
		$M_{c, s}$                 & maximum number of blocks clinician $c$ should
		cover on service $s$ \\
    \bottomrule
	\end{tabular}
\end{table}

\subsection{Constraints}\label{sec:meth-constraints}
We now formalize the hard constraints in Table~\ref{tbl:constraint-summary}
using the variables defined above. 
The BC (block coverage) and WC (weekend coverage) constraints, 
are given by Eqns. (\ref{eqn:constr-block-cov}) and (\ref{eqn:constr-weekend-cov}), respectively. 
\begin{align}
&\sum_{c=1}^{C} X_{c, b, s} = 1 &&\forall b\in \mathcal{B}, s \in \mathcal{S} \label{eqn:constr-block-cov} \\
&\sum_{c=1}^{C} Y_{c, w} = 1 &&\forall w\in \mathcal{W} \label{eqn:constr-weekend-cov}
\end{align}
The MM (min/max) constraint is given by:
\begin{align}
&m_{c, s} \leq \sum_{b=1}^{B} X_{c, b, s} \leq M_{c, s} &&\forall\
c\in\mathcal{C}, s\in\mathcal{S} \label{eqn:constr-min-max}
\end{align}
The NCB (no consecutive blocks) and NCW (no consecutive weekends) constraints are 
given by Eqns. (\ref{eqn:constr-no-consec-blocks}) and (\ref{eqn:constr-no-consec-weekends}), respectively.
\begin{align}
&X_{c, b, s} + X_{c, b + 1, s} \leq 1 &&\forall c\in\mathcal{C}, b \leq B - 1,
s\in\mathcal{S} \label{eqn:constr-no-consec-blocks} \\
&Y_{c, w} + Y_{c, w + 1} \leq 1 &&\forall c\in\mathcal{C}, w \leq W - 1 \label{eqn:constr-no-consec-weekends}
\end{align}
The EW (equal weekend) and EH (equal holidays) constraints are given by Eqns. (\ref{eqn:constr-equal-weekends})
and (\ref{eqn:constr-equal-holidays}), respectively.
\begin{align}
&\floor*{\frac{W}{C}} \leq \sum_{w=1}^W Y_{c, w} \leq \ceil*{\frac{W}{C}}
&&\forall c\in\mathcal{C} \label{eqn:constr-equal-weekends} \\
&\floor*{\frac{\abs{\mathcal{L}}}{C}} \leq \sum_{w\in\mathcal{L}} Y_{c, w} \leq
\ceil*{\frac{\abs{\mathcal{L}}}{C}} &&\forall c\in\mathcal{C}
\label{eqn:constr-equal-holidays}
\end{align}

\subsection{Objectives}\label{sec:meth-objectives}
As described in Section~\ref{sec:problem}, the soft constraints of the clinician
scheduling problem include: satisfying clinician block off requests (BR),
satisfying clinician weekend off requests (WR), and assigning weekends closer to
blocks (BWA). We convert these soft constraints into linear objective functions
of the binary variables defined in Section~\ref{sec:meth-variables}. 
Objectives BR and WR are given in Eqns. (\ref{eqn:obj-block-requests}) and (\ref{eqn:obj-weekend-requests})
as linear functions of $X$ and $Y$:
\begin{align}
&Q_1(X) = \sum_{c=1}^{C} \sum_{b=1}^{B} \sum_{s=1}^{S}
{(-1)}^{\ind(b\,\in\,\mathcal{U}_c)}\cdot X_{c, b, s}
\label{eqn:obj-block-requests}\\
&Q_2(Y) = \sum_{c=1}^{C} \sum_{w=1}^{W}
{(-1)}^{\ind(w\,\in\,\mathcal{V}_c)}\cdot Y_{c, w}
\label{eqn:obj-weekend-requests}
\end{align}
where $\ind(P)$ is the indicator function that has value 1 when predicate $P$
holds and 0 otherwise. In the above two objectives, we penalize any assignments
that conflict with a block or weekend request, and aim to maximize the
non-conflicting assignments.

The BWA objective is optimized by considering the product $X_{c, b, s}\cdot Y_{c, w}$ 
for values of $w$ ``adjacent'' to the value of $b$. This
leads to the maximization objective:
\begin{align}
&Q_3(X, Y) = \sum_{c=1}^{C} \sum_{b=1}^{B} \sum_{s=1}^{S} X_{c, b, s}\cdot Y_{c,
	w=\varphi(b)} \label{eqn:obj-block-weekend-adj}
\end{align}
where $\varphi(b)$ is a one-to-one mapping of a block to an adjacent weekend, by
some appropriate definition of adjacency. For instance, clinicians might want to
be assigned during a weekend that falls within an assigned block. In this case,
we will have $\varphi(b) = 2b-1$.

However, as it is, $Q_3$ is not a linear function of the assignment variables
$X$ and $Y$, and cannot be optimized in a linear programming framework. 
An approach used to convert such
functions into linear objectives involves introducing a helper variable and
additional constraints~\cite{hammer_boolean_1968}. 
In our case, we introduce a variable $Z_{c, b, s}$ 
for every product $X_{c, b, s} \cdot Y_{c, w}$ with $w = \varphi(b)$, and
constraining $Z$ such that 
\begin{align}
&Z_{c, b, s} \leq X_{c, b, s} \label{eqn:helper-x-constraint}\\
&Z_{c, b, s} \leq Y_{c, w=\varphi(b)} &&\forall s\in\mathcal{S}
\label{eqn:helper-y-constraint}
\end{align}
Since $X_{c, b, s}$ and $Y_{c, w}$ are binary variables, $Z_{c, b, s}$ will be constrained
to 0, unless both $X_{c, b, s}$ and $Y_{c, w}$ are 1. Therefore, it suffices to maximize
the following linear function of $Z$,
\begin{align}
&Q_3(Z) = \sum_{c=1}^{C} \sum_{b=1}^{B} \sum_{s=1}^{S} Z_{c, b, s}
\end{align}
to get the correct adjacency maximization objective.

In order to optimize all objectives simultaneously, we optimize a weighted sum
of the normalized objective functions,
\begin{equation}
\max_{X, Y, Z} \alpha_1 \bar{Q}_1(X) + \alpha_2 \bar{Q}_2(Y) + \alpha_3
\bar{Q}_3(Z)
\end{equation}
subject to the constraints defined in Section \ref{sec:meth-constraints},
where $\bar{Q}_i$ is the normalization of objective $Q_i$, and $0 \leq \alpha_i \leq 1$. 
This method guarantees an optimal solution to be Pareto optimal~\cite{stanimirovic_linear_2011}. 

Currently, the most efficient approach to finding an exact solution for 
an ILP is called Branch-and-Cut~\cite{mitchell_branch-and-cut_2002}.
This method involves iteratively solving 
LP relaxations of the ILP, then constraining the relaxed problems and
considering various sub-problems until it finds integral solutions to the 
original ILP.
In the intermediate relaxations, the integer assignment variables can
take on real values, allowing the problem to be solved efficiently using 
the Simplex method~\cite{shamir_efficiency_1987}. The 
complexity of finding an optimal integral solution thus lies in the 
branching search structure of Branch-and-Cut.
	\section{Experiments}\label{sec:experiments}
	Our goal was to determine if the schedules provided by solving the ILP 
could successfully
(i) enforce all hard constraints; 
(ii) improve fulfillment of soft constraints compared to the manual approach; 
and (iii) assess whether our ILP formulation can be used for a wide range
of configurations.
First, we compared the schedules created by solving the ILP formulation
given in Section~\ref{sec:methods} to schedules that were manually generated,
with respect to adherence to the hard and soft constraints outlined in Section~\ref{sec:problem}.
We then examined the efficiency of the ILP approach in generating schedules
by its runtime on a variety of instances that may be found in the real-world.

\subsection{Implementation}
We developed a Python software package with a user interface that implements the above
linear program and allows configuration of clinicians, to be
used by the ID division at St.\ Michael's Hospital~\cite{landsman_scheduling}. 
The software was used to
generate the results in the following sections, using real data as well as
simulated data as input. All the following experiments were conducted on an
Intel Core i7-4770k CPU @ 3.50 GHz with 16 GB of RAM running 64-bit Windows 10.
Our software package uses COIN-OR Branch-and-Cut open source solver
version 2.9.9~\cite{johnjforrest_coin-or/cbc:_2019}.

\subsection{Comparison with Manually Generated Schedules}
We used clinician time-off requests and minimum/maximum requirements from
2015-2018 as input data for the ILP problem.
Table~\ref{tbl:2018-schedule-comparison} compares the optimal schedule generated using
the software with the manually-created schedule for data from 2018. The
schedule is color-coded to distinguish between the different clinicians.

\begin{table}[htbp]
  \centering
  \caption{Comparison of automatically generated (ILP solution) and manually generated schedules for 2018. In the ILP solution there are rigid 2-week block assignments, unlike manual generation that often assigns 3-4 weeks in a row to a single clinician. Moreover, in the ILP solution we see for each block either the HIV or ID clinician was assigned weekend coverage, indicating improved Block-Weekend Adjacency.}%
  \label{tbl:2018-schedule-comparison}%
	\begin{adjustbox}{scale=0.8}
    \begin{tabular}{c||ccc||ccc}
    \multicolumn{1}{c||}{\multirow{2}[1]{*}{Week \#}} & \multicolumn{3}{c||}{ILP Solution} & \multicolumn{3}{c}{Manual Generation} \\
          & HIV   & ID    & Weekend & HIV   & ID    & Weekend \\
    \midrule
    \midrule
    1     & \cellcolor[rgb]{ .663,  .816,  .557}A & \cellcolor[rgb]{ .957,  .69,  .518}E & \cellcolor[rgb]{ .957,  .69,  .518}E & \cellcolor[rgb]{ .663,  .816,  .557}A & \cellcolor[rgb]{ .957,  .69,  .518}E & \cellcolor[rgb]{ .459,  .443,  .443}H \\
    2     & \cellcolor[rgb]{ .663,  .816,  .557}A & \cellcolor[rgb]{ .957,  .69,  .518}E & \cellcolor[rgb]{ .459,  .443,  .443}H & \cellcolor[rgb]{ .663,  .816,  .557}A & \cellcolor[rgb]{ .957,  .69,  .518}E & \cellcolor[rgb]{ .663,  .816,  .557}A \\
    3     & \cellcolor[rgb]{ .608,  .761,  .902}B & \cellcolor[rgb]{ .557,  .663,  .859}F & \cellcolor[rgb]{ .557,  .663,  .859}F & \cellcolor[rgb]{ .608,  .761,  .902}B & \cellcolor[rgb]{ .459,  .443,  .443}H & \cellcolor[rgb]{ .518,  .592,  .69}G \\
    4     & \cellcolor[rgb]{ .608,  .761,  .902}B & \cellcolor[rgb]{ .557,  .663,  .859}F & \cellcolor[rgb]{ .459,  .443,  .443}H & \cellcolor[rgb]{ .608,  .761,  .902}B & \cellcolor[rgb]{ .459,  .443,  .443}H & \cellcolor[rgb]{ .251,  .251,  .251}\textcolor[rgb]{ 1,  1,  1}{I} \\
    5     & \cellcolor[rgb]{ .663,  .816,  .557}A & \cellcolor[rgb]{ .518,  .592,  .69}G & \cellcolor[rgb]{ .663,  .816,  .557}A & \cellcolor[rgb]{ .663,  .816,  .557}A & \cellcolor[rgb]{ .518,  .592,  .69}G & \cellcolor[rgb]{ .557,  .663,  .859}F \\
    6     & \cellcolor[rgb]{ .663,  .816,  .557}A & \cellcolor[rgb]{ .518,  .592,  .69}G & \cellcolor[rgb]{ .957,  .69,  .518}E & \cellcolor[rgb]{ .663,  .816,  .557}A & \cellcolor[rgb]{ .518,  .592,  .69}G & \cellcolor[rgb]{ 1,  .851,  .4}C \\
    7     & \cellcolor[rgb]{ .608,  .761,  .902}B & \cellcolor[rgb]{ 1,  .851,  .4}C & \cellcolor[rgb]{ 1,  .851,  .4}C & \cellcolor[rgb]{ .663,  .816,  .557}A & \cellcolor[rgb]{ .557,  .663,  .859}F & \cellcolor[rgb]{ .608,  .761,  .902}B \\
    8     & \cellcolor[rgb]{ .608,  .761,  .902}B & \cellcolor[rgb]{ 1,  .851,  .4}C & \cellcolor[rgb]{ .518,  .592,  .69}G & \cellcolor[rgb]{ .788,  .788,  .788}D & \cellcolor[rgb]{ 1,  .851,  .4}C & \cellcolor[rgb]{ .518,  .592,  .69}G \\
    9     & \cellcolor[rgb]{ .788,  .788,  .788}D & \cellcolor[rgb]{ .251,  .251,  .251}\textcolor[rgb]{ 1,  1,  1}{I} & \cellcolor[rgb]{ .788,  .788,  .788}D & \cellcolor[rgb]{ .608,  .761,  .902}B & \cellcolor[rgb]{ 1,  .851,  .4}C & \cellcolor[rgb]{ .788,  .788,  .788}D \\
    10    & \cellcolor[rgb]{ .788,  .788,  .788}D & \cellcolor[rgb]{ .251,  .251,  .251}\textcolor[rgb]{ 1,  1,  1}{I} & \cellcolor[rgb]{ .459,  .443,  .443}H & \cellcolor[rgb]{ .608,  .761,  .902}B & \cellcolor[rgb]{ .788,  .788,  .788}D & \cellcolor[rgb]{ .459,  .443,  .443}H \\
    11    & \cellcolor[rgb]{ .663,  .816,  .557}A & \cellcolor[rgb]{ .608,  .761,  .902}B & \cellcolor[rgb]{ .608,  .761,  .902}B & \cellcolor[rgb]{ .663,  .816,  .557}A & \cellcolor[rgb]{ .608,  .761,  .902}B & \cellcolor[rgb]{ .557,  .663,  .859}F \\
    12    & \cellcolor[rgb]{ .663,  .816,  .557}A & \cellcolor[rgb]{ .608,  .761,  .902}B & \cellcolor[rgb]{ .251,  .251,  .251}\textcolor[rgb]{ 1,  1,  1}{I} & \cellcolor[rgb]{ .663,  .816,  .557}A & \cellcolor[rgb]{ .608,  .761,  .902}B & \cellcolor[rgb]{ .663,  .816,  .557}A \\
    13    & \cellcolor[rgb]{ 1,  .851,  .4}C & \cellcolor[rgb]{ .557,  .663,  .859}F & \cellcolor[rgb]{ .557,  .663,  .859}F & \cellcolor[rgb]{ 1,  .851,  .4}C & \cellcolor[rgb]{ .459,  .443,  .443}H & \cellcolor[rgb]{ .459,  .443,  .443}H \\
    14    & \cellcolor[rgb]{ 1,  .851,  .4}C & \cellcolor[rgb]{ .557,  .663,  .859}F & \cellcolor[rgb]{ .251,  .251,  .251}\textcolor[rgb]{ 1,  1,  1}{I} & \cellcolor[rgb]{ 1,  .851,  .4}C & \cellcolor[rgb]{ .459,  .443,  .443}H & \cellcolor[rgb]{ .251,  .251,  .251}\textcolor[rgb]{ 1,  1,  1}{I} \\
    15    & \cellcolor[rgb]{ .663,  .816,  .557}A & \cellcolor[rgb]{ .459,  .443,  .443}H & \cellcolor[rgb]{ .459,  .443,  .443}H & \cellcolor[rgb]{ .608,  .761,  .902}B & \cellcolor[rgb]{ .251,  .251,  .251}\textcolor[rgb]{ 1,  1,  1}{I} & \cellcolor[rgb]{ 1,  .851,  .4}C \\
    16    & \cellcolor[rgb]{ .663,  .816,  .557}A & \cellcolor[rgb]{ .459,  .443,  .443}H & \cellcolor[rgb]{ .788,  .788,  .788}D & \cellcolor[rgb]{ .608,  .761,  .902}B & \cellcolor[rgb]{ .251,  .251,  .251}\textcolor[rgb]{ 1,  1,  1}{I} & \cellcolor[rgb]{ .957,  .69,  .518}E \\
    17    & \cellcolor[rgb]{ .608,  .761,  .902}B & \cellcolor[rgb]{ .957,  .69,  .518}E & \cellcolor[rgb]{ .608,  .761,  .902}B & \cellcolor[rgb]{ .663,  .816,  .557}A & \cellcolor[rgb]{ .957,  .69,  .518}E & \cellcolor[rgb]{ .788,  .788,  .788}D \\
    18    & \cellcolor[rgb]{ .608,  .761,  .902}B & \cellcolor[rgb]{ .957,  .69,  .518}E & \cellcolor[rgb]{ .459,  .443,  .443}H & \cellcolor[rgb]{ .663,  .816,  .557}A & \cellcolor[rgb]{ .957,  .69,  .518}E & \cellcolor[rgb]{ .957,  .69,  .518}E \\
    19    & \cellcolor[rgb]{ .663,  .816,  .557}A & \cellcolor[rgb]{ .251,  .251,  .251}\textcolor[rgb]{ 1,  1,  1}{I} & \cellcolor[rgb]{ .251,  .251,  .251}\textcolor[rgb]{ 1,  1,  1}{I} & \cellcolor[rgb]{ .663,  .816,  .557}A & \cellcolor[rgb]{ 1,  .851,  .4}C & \cellcolor[rgb]{ .557,  .663,  .859}F \\
    20    & \cellcolor[rgb]{ .663,  .816,  .557}A & \cellcolor[rgb]{ .251,  .251,  .251}\textcolor[rgb]{ 1,  1,  1}{I} & \cellcolor[rgb]{ .788,  .788,  .788}D & \cellcolor[rgb]{ .663,  .816,  .557}A & \cellcolor[rgb]{ 1,  .851,  .4}C & \cellcolor[rgb]{ 1,  .851,  .4}C \\
    21    & \cellcolor[rgb]{ 1,  .851,  .4}C & \cellcolor[rgb]{ .788,  .788,  .788}D & \cellcolor[rgb]{ 1,  .851,  .4}C & \cellcolor[rgb]{ .608,  .761,  .902}B & \cellcolor[rgb]{ .518,  .592,  .69}G & \cellcolor[rgb]{ .663,  .816,  .557}A \\
    22    & \cellcolor[rgb]{ 1,  .851,  .4}C & \cellcolor[rgb]{ .788,  .788,  .788}D & \cellcolor[rgb]{ .251,  .251,  .251}\textcolor[rgb]{ 1,  1,  1}{I} & \cellcolor[rgb]{ .608,  .761,  .902}B & \cellcolor[rgb]{ .518,  .592,  .69}G & \cellcolor[rgb]{ 1,  .851,  .4}C \\
    23    & \cellcolor[rgb]{ .608,  .761,  .902}B & \cellcolor[rgb]{ .557,  .663,  .859}F & \cellcolor[rgb]{ .557,  .663,  .859}F & \cellcolor[rgb]{ 1,  .851,  .4}C & \cellcolor[rgb]{ .557,  .663,  .859}F & \cellcolor[rgb]{ .788,  .788,  .788}D \\
    24    & \cellcolor[rgb]{ .608,  .761,  .902}B & \cellcolor[rgb]{ .557,  .663,  .859}F & \cellcolor[rgb]{ .518,  .592,  .69}G & \cellcolor[rgb]{ 1,  .851,  .4}C & \cellcolor[rgb]{ .557,  .663,  .859}F & \cellcolor[rgb]{ 1,  .851,  .4}C \\
    25    & \cellcolor[rgb]{ .663,  .816,  .557}A & \cellcolor[rgb]{ .459,  .443,  .443}H & \cellcolor[rgb]{ .663,  .816,  .557}A & \cellcolor[rgb]{ 1,  .851,  .4}C & \cellcolor[rgb]{ 1,  .851,  .4}C & \cellcolor[rgb]{ .518,  .592,  .69}G \\
    26    & \cellcolor[rgb]{ .663,  .816,  .557}A & \cellcolor[rgb]{ .459,  .443,  .443}H & \cellcolor[rgb]{ .459,  .443,  .443}H & \cellcolor[rgb]{ .788,  .788,  .788}D & \cellcolor[rgb]{ .251,  .251,  .251}\textcolor[rgb]{ 1,  1,  1}{I} & \cellcolor[rgb]{ .788,  .788,  .788}D \\
    27    & \cellcolor[rgb]{ 1,  .851,  .4}C & \cellcolor[rgb]{ .788,  .788,  .788}D & \cellcolor[rgb]{ .788,  .788,  .788}D & \cellcolor[rgb]{ .663,  .816,  .557}A & \cellcolor[rgb]{ .608,  .761,  .902}B & \cellcolor[rgb]{ .957,  .69,  .518}E \\
    28    & \cellcolor[rgb]{ 1,  .851,  .4}C & \cellcolor[rgb]{ .788,  .788,  .788}D & \cellcolor[rgb]{ .957,  .69,  .518}E & \cellcolor[rgb]{ .663,  .816,  .557}A & \cellcolor[rgb]{ .608,  .761,  .902}B & \cellcolor[rgb]{ .251,  .251,  .251}\textcolor[rgb]{ 1,  1,  1}{I} \\
    29    & \cellcolor[rgb]{ .663,  .816,  .557}A & \cellcolor[rgb]{ .608,  .761,  .902}B & \cellcolor[rgb]{ .663,  .816,  .557}A & \cellcolor[rgb]{ .608,  .761,  .902}B & \cellcolor[rgb]{ .788,  .788,  .788}D & \cellcolor[rgb]{ .788,  .788,  .788}D \\
    30    & \cellcolor[rgb]{ .663,  .816,  .557}A & \cellcolor[rgb]{ .608,  .761,  .902}B & \cellcolor[rgb]{ .608,  .761,  .902}B & \cellcolor[rgb]{ .608,  .761,  .902}B & \cellcolor[rgb]{ .788,  .788,  .788}D & \cellcolor[rgb]{ .663,  .816,  .557}A \\
    31    & \cellcolor[rgb]{ 1,  .851,  .4}C & \cellcolor[rgb]{ .957,  .69,  .518}E & \cellcolor[rgb]{ 1,  .851,  .4}C & \cellcolor[rgb]{ 1,  .851,  .4}C & \cellcolor[rgb]{ .557,  .663,  .859}F & \cellcolor[rgb]{ .957,  .69,  .518}E \\
    32    & \cellcolor[rgb]{ 1,  .851,  .4}C & \cellcolor[rgb]{ .957,  .69,  .518}E & \cellcolor[rgb]{ .663,  .816,  .557}A & \cellcolor[rgb]{ 1,  .851,  .4}C & \cellcolor[rgb]{ .557,  .663,  .859}F & \cellcolor[rgb]{ .557,  .663,  .859}F \\
    33    & \cellcolor[rgb]{ .608,  .761,  .902}B & \cellcolor[rgb]{ .788,  .788,  .788}D & \cellcolor[rgb]{ .788,  .788,  .788}D & \cellcolor[rgb]{ .608,  .761,  .902}B & \cellcolor[rgb]{ .557,  .663,  .859}F & \cellcolor[rgb]{ .251,  .251,  .251}\textcolor[rgb]{ 1,  1,  1}{I} \\
    34    & \cellcolor[rgb]{ .608,  .761,  .902}B & \cellcolor[rgb]{ .788,  .788,  .788}D & \cellcolor[rgb]{ .957,  .69,  .518}E & \cellcolor[rgb]{ .608,  .761,  .902}B & \cellcolor[rgb]{ .251,  .251,  .251}\textcolor[rgb]{ 1,  1,  1}{I} & \cellcolor[rgb]{ 1,  .851,  .4}C \\
    35    & \cellcolor[rgb]{ .663,  .816,  .557}A & \cellcolor[rgb]{ .251,  .251,  .251}\textcolor[rgb]{ 1,  1,  1}{I} & \cellcolor[rgb]{ .663,  .816,  .557}A & \cellcolor[rgb]{ .663,  .816,  .557}A & \cellcolor[rgb]{ .518,  .592,  .69}G & \cellcolor[rgb]{ .518,  .592,  .69}G \\
    36    & \cellcolor[rgb]{ .663,  .816,  .557}A & \cellcolor[rgb]{ .251,  .251,  .251}\textcolor[rgb]{ 1,  1,  1}{I} & \cellcolor[rgb]{ .518,  .592,  .69}G & \cellcolor[rgb]{ .663,  .816,  .557}A & \cellcolor[rgb]{ .518,  .592,  .69}G & \cellcolor[rgb]{ .251,  .251,  .251}\textcolor[rgb]{ 1,  1,  1}{I} \\
    37    & \cellcolor[rgb]{ .788,  .788,  .788}D & \cellcolor[rgb]{ 1,  .851,  .4}C & \cellcolor[rgb]{ 1,  .851,  .4}C & \cellcolor[rgb]{ .788,  .788,  .788}D & \cellcolor[rgb]{ 1,  .851,  .4}C & \cellcolor[rgb]{ .663,  .816,  .557}A \\
    38    & \cellcolor[rgb]{ .788,  .788,  .788}D & \cellcolor[rgb]{ 1,  .851,  .4}C & \cellcolor[rgb]{ .557,  .663,  .859}F & \cellcolor[rgb]{ .788,  .788,  .788}D & \cellcolor[rgb]{ .788,  .788,  .788}D & \cellcolor[rgb]{ .957,  .69,  .518}E \\
    39    & \cellcolor[rgb]{ .608,  .761,  .902}B & \cellcolor[rgb]{ .957,  .69,  .518}E & \cellcolor[rgb]{ .957,  .69,  .518}E & \cellcolor[rgb]{ .663,  .816,  .557}A & \cellcolor[rgb]{ .608,  .761,  .902}B & \cellcolor[rgb]{ .788,  .788,  .788}D \\
    40    & \cellcolor[rgb]{ .608,  .761,  .902}B & \cellcolor[rgb]{ .957,  .69,  .518}E & \cellcolor[rgb]{ .608,  .761,  .902}B & \cellcolor[rgb]{ .608,  .761,  .902}B & \cellcolor[rgb]{ .251,  .251,  .251}\textcolor[rgb]{ 1,  1,  1}{I} & \cellcolor[rgb]{ .251,  .251,  .251}\textcolor[rgb]{ 1,  1,  1}{I} \\
    41    & \cellcolor[rgb]{ .663,  .816,  .557}A & \cellcolor[rgb]{ .518,  .592,  .69}G & \cellcolor[rgb]{ .518,  .592,  .69}G & \cellcolor[rgb]{ .608,  .761,  .902}B & \cellcolor[rgb]{ .251,  .251,  .251}\textcolor[rgb]{ 1,  1,  1}{I} & \cellcolor[rgb]{ .518,  .592,  .69}G \\
    42    & \cellcolor[rgb]{ .663,  .816,  .557}A & \cellcolor[rgb]{ .518,  .592,  .69}G & \cellcolor[rgb]{ .957,  .69,  .518}E & \cellcolor[rgb]{ .788,  .788,  .788}D & \cellcolor[rgb]{ .557,  .663,  .859}F & \cellcolor[rgb]{ .557,  .663,  .859}F \\
    43    & \cellcolor[rgb]{ 1,  .851,  .4}C & \cellcolor[rgb]{ .251,  .251,  .251}\textcolor[rgb]{ 1,  1,  1}{I} & \cellcolor[rgb]{ 1,  .851,  .4}C & \cellcolor[rgb]{ 1,  .851,  .4}C & \cellcolor[rgb]{ .557,  .663,  .859}F & \cellcolor[rgb]{ 1,  .851,  .4}C \\
    44    & \cellcolor[rgb]{ 1,  .851,  .4}C & \cellcolor[rgb]{ .251,  .251,  .251}\textcolor[rgb]{ 1,  1,  1}{I} & \cellcolor[rgb]{ .251,  .251,  .251}\textcolor[rgb]{ 1,  1,  1}{I} & \cellcolor[rgb]{ .788,  .788,  .788}D & \cellcolor[rgb]{ .957,  .69,  .518}E & \cellcolor[rgb]{ .251,  .251,  .251}\textcolor[rgb]{ 1,  1,  1}{I} \\
    45    & \cellcolor[rgb]{ .663,  .816,  .557}A & \cellcolor[rgb]{ .557,  .663,  .859}F & \cellcolor[rgb]{ .663,  .816,  .557}A & \cellcolor[rgb]{ .788,  .788,  .788}D & \cellcolor[rgb]{ .957,  .69,  .518}E & \cellcolor[rgb]{ .957,  .69,  .518}E \\
    46    & \cellcolor[rgb]{ .663,  .816,  .557}A & \cellcolor[rgb]{ .557,  .663,  .859}F & \cellcolor[rgb]{ .557,  .663,  .859}F & \cellcolor[rgb]{ .663,  .816,  .557}A & \cellcolor[rgb]{ .608,  .761,  .902}B & \cellcolor[rgb]{ .663,  .816,  .557}A \\
    47    & \cellcolor[rgb]{ .608,  .761,  .902}B & \cellcolor[rgb]{ .518,  .592,  .69}G & \cellcolor[rgb]{ .608,  .761,  .902}B & \cellcolor[rgb]{ .663,  .816,  .557}A & \cellcolor[rgb]{ .608,  .761,  .902}B & \cellcolor[rgb]{ .788,  .788,  .788}D \\
    48    & \cellcolor[rgb]{ .608,  .761,  .902}B & \cellcolor[rgb]{ .518,  .592,  .69}G & \cellcolor[rgb]{ .251,  .251,  .251}\textcolor[rgb]{ 1,  1,  1}{I} & \cellcolor[rgb]{ .663,  .816,  .557}A & \cellcolor[rgb]{ .788,  .788,  .788}D & \cellcolor[rgb]{ .518,  .592,  .69}G \\
    49    & \cellcolor[rgb]{ .788,  .788,  .788}D & \cellcolor[rgb]{ 1,  .851,  .4}C & \cellcolor[rgb]{ .788,  .788,  .788}D & \cellcolor[rgb]{ .663,  .816,  .557}A & \cellcolor[rgb]{ .788,  .788,  .788}D & \cellcolor[rgb]{ .557,  .663,  .859}F \\
    50    & \cellcolor[rgb]{ .788,  .788,  .788}D & \cellcolor[rgb]{ 1,  .851,  .4}C & \cellcolor[rgb]{ .608,  .761,  .902}B & \cellcolor[rgb]{ .608,  .761,  .902}B & \cellcolor[rgb]{ .518,  .592,  .69}G & \cellcolor[rgb]{ .518,  .592,  .69}G \\
    51    & \cellcolor[rgb]{ .608,  .761,  .902}B & \cellcolor[rgb]{ .518,  .592,  .69}G & \cellcolor[rgb]{ .518,  .592,  .69}G & \cellcolor[rgb]{ .608,  .761,  .902}B & \cellcolor[rgb]{ .518,  .592,  .69}G & \cellcolor[rgb]{ .957,  .69,  .518}E \\
    \end{tabular}%
	\end{adjustbox}\\[1em]
	\footnotesize\raggedright
	The ``HIV/ID'' columns represent the assignments for the two concurrent services offered at the department. The ``Weekend'' column represents the assignments for weekend coverage in both services. Different colours and letters are used to distinguish different clinicians.
\end{table}%

First, we evaluated the ILP solution by comparing it with the
manual generation, as in Table~\ref{tbl:2018-schedule-comparison}. Specifically, we examined the adherence of each
schedule to the constraints presented in Table~\ref{tbl:constraint-summary}. As
shown in Table~\ref{tbl:constraints-comparison}, the ILP solution satisfied all 
hard constraints. In contrast,
manual generation did not satisfy all hard constraints. In
particular, we see that the manual generation assigned clinicians to multiple
consecutive blocks in all four years. Moreover, the manual generation did not have
an equal distribution of weekends and holidays for all four years of data.
Considering all objectives, we see that the ILP solution outperforms manual generation
in all four years, by accommodating almost all time-off requests and
ensuring that weekends are always assigned close to blocks.

\begin{sidewaystable}[htbp]
    \centering
    \caption{Comparison of constraint satisfaction and objectives values in LP-generated and historical schedules.}%
    \label{tbl:constraints-comparison}
		\begin{tabular}{l|cc|cc|cc|cc}
      \toprule
			\multirow{2}[1]{*}{}                                & \multicolumn{2}{c|}{\textbf{2015}} & \multicolumn{2}{c|}{\textbf{2016}} & \multicolumn{2}{c|}{\textbf{2017}} & \multicolumn{2}{c}{\textbf{2018}} \\
			                                                    &     LP     &      Historical       &     LP     &      Historical       &     LP     &      Historical       &     LP     &      Historical      \\ \midrule
			\multicolumn{1}{c|}{\textbf{Constraint}}            &            &                       &            &                       &            &                       &            &                      \\ \midrule
			Block Coverage                                      & \checkmark &      \checkmark       & \checkmark &      \checkmark       & \checkmark &      \checkmark       & \checkmark &      \checkmark      \\
			Weekend Coverage                                    & \checkmark &      \checkmark       & \checkmark &      \checkmark       & \checkmark &      \checkmark       & \checkmark &      \checkmark      \\
			Min/Max                                             & \checkmark &      \checkmark       & \checkmark &      \checkmark       & \checkmark &      \checkmark       & \checkmark &      \checkmark      \\
			No Consecutive Blocks                               & \checkmark &                       & \checkmark &                       & \checkmark &                       & \checkmark &                      \\
			No Consecutive Weekends                             & \checkmark &                       & \checkmark &                       & \checkmark &      \checkmark       & \checkmark &      \checkmark      \\
			Equal Weekends                                      & \checkmark &                       & \checkmark &                       & \checkmark &                       & \checkmark &                      \\
			Equal Holidays                                      & \checkmark &                       & \checkmark &      \checkmark       & \checkmark &                       & \checkmark &      \checkmark      \\ \midrule
			\multicolumn{1}{c|}{\textbf{Objective}}             &            &                       &            &                       &            &                       &            &                      \\ \midrule
			\makecell[l]{Satisfied Block Requests}              &  123/129   &        121/129        &  120/126   &        116/126        &   99/99    &         95/99         &  124/128   &       121/128        \\
			\makecell[l]{Satisfied Weekend Requests}            &  113/113   &        111/113        &  119/119   &        112/119        &   75/75    &         75/75         &  115/115   &       113/115        \\
			\makecell[l]{Adjacent Block-Weekend Assignments}    &   26/26    &         9/26          &   26/26    &         6/26          &   26/26    &         7/26          &   26/26    &         5/26         \\
      \bottomrule
		\end{tabular}\\[1em]
  \footnotesize\raggedright
  The objective denominator represents:
  the total number of requests submitted by clinicians for request objectives; and
  the total number of possible adjacencies for adjacency objectives.
\end{sidewaystable}%

\subsection{Influence of Problem Complexity on Runtime}
Next, we examined the influence of the following four parameters on the
runtime of the ILP solver using simulated data:
number of clinicians;
number of services offered;
number of time-off requests per clinician per year;
time-horizon of the schedule.

The effect of increasing the number of clinicians and number of services 
on the runtime of the program is shown in Table~\ref{tbl:runtime-services-clinicians-comparison}.
We executed the algorithm for
$S = \{1, 2, 3\}$ total services and $C = \{10, 20, 30, 50\}$ clinicians in  
total across all services. 
In a department providing a single service, increasing the number of clinicians
did not affect the runtime, and we were able to find an ILP solution in all four cases
within 1 second.
For 2 concurrent services, a roster of 30 or more
clinicians becomes impractical to schedule, as searching for a solution required
over 24 hours. We saw similar issues for a roster of 20 or more clinicians
assigned to a division with 3 concurrent services. However, when removing the
NCB constraint, we saw a great improvement in runtime for divisions with 2 and 3
services, and we were able to generate a schedule with upwards of 50 clinicians
in under 1.5 seconds.

\begin{table}[htbp]
	\centering
 	\caption{Comparison of program runtime (in seconds) for different numbers of services and total clinicians in the division.}%
  \label{tbl:runtime-services-clinicians-comparison}%
	\begin{tabular}{|c|c||c|c||c|c|}
		\toprule
		                                      &  \multicolumn{5}{c|}{Number of Services}  \\ \midrule
		\makecell[l]{Number of \\ Clinicians} &  1   &    2    & 2 (*) &  3   & 3 (*) \\ \midrule
		                 10                   & 0.16 &  0.74   &  0.16   & 1.40 &  0.23   \\ \hline
		                 20                   & 0.25 & 7468.86 &  0.32   &  --  &  0.43   \\ \hline
		                 30                   & 0.42 &    --   &  0.49   &  --  &  0.66   \\ \hline
		                 50                   & 0.62 &    --   &  0.82   &  --  &  1.14   \\ \bottomrule
	\end{tabular}\\[1em]
  \footnotesize\raggedright
  Notes:
  ``--'' indicates that no solution was found within 24 hours;
  ``(*)'' indicates that the No Consecutive Blocks (NCB) constraint was removed.
\end{table}

\begin{figure}[h]
	\centering
	\def\svgwidth{\columnwidth}
	\caption{Runtime of ILP solver with an increasing number of requests per clinician}
	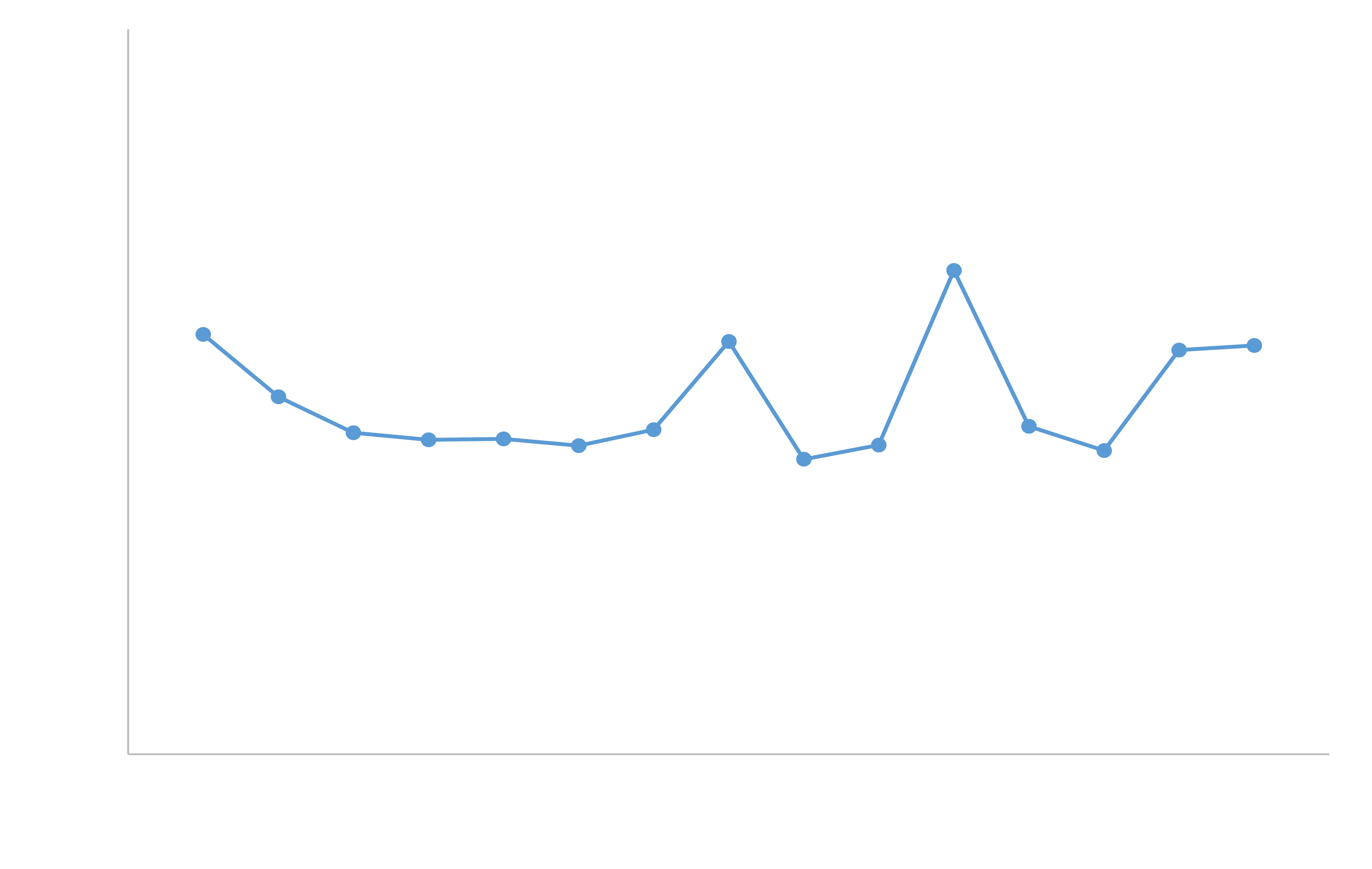
  \label{fig:runtime-requests}
\end{figure}

For the remaining experiments, we simulated a department with 10 clinicians offering 
two services, similar to the department at St.\ Michael's Hospital. 
The effect of an increasing number of requests per clinician on the runtime of
the ILP solver is shown in Figure~\ref{fig:runtime-requests}.
In this experiment, each clinician was configured with 1 to 15 total block requests.
The runtime of the algorithm is constant with respect to the number of requests,
indicating that it can accommodate a lot of flexibility in
clinician requests. Moreover, we see that all runs were completed in under
2 seconds.

\begin{figure}[h]
	\centering
	\def\svgwidth{\textwidth}
	\caption{Runtime of ILP solver with an increasing number of 2-week blocks}%
	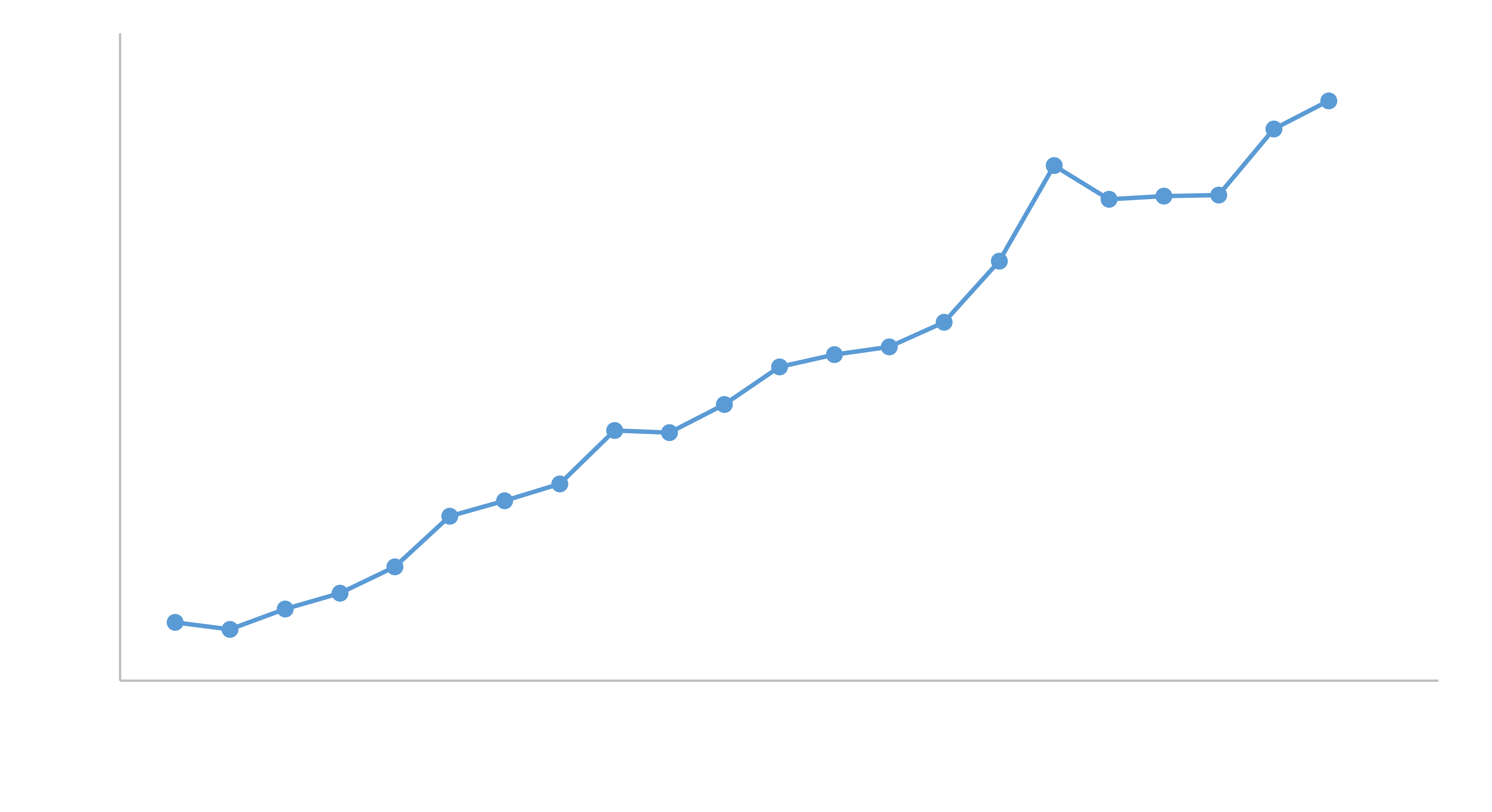
  \label{fig:runtime-blocks}
\end{figure}

Figure~\ref{fig:runtime-blocks} presents the change in runtime when increasing
the number of 2-week blocks in a department with 10
clinicians offering two services. In this experiment, we investigated time horizons
from 5 to 110 blocks. This is equivalent to generating a schedule for up to 4 years ahead.
The trend in the graph indicates a linear growth in runtime with respect to the time-horizon.
Notably, the ILP solver was able to find all solutions in under 6 seconds, indicating very 
good performance for long-term scheduling. 

	\section{Discussion}\label{sec:discussion}
	In this paper, we present a simple, yet flexible, integer linear programming
formulation to generate schedules for clinical departments at hospitals.
The challenge in applying ILP to the task of scheduling clinicians lies in the
computational complexity of finding an optimal solution. As the size of the
scheduling problem grows, due to a larger roster of clinicians or more
complicated constraints, the time it takes to generate an optimal schedule may
grow exponentially in the worst case.
Many previous approaches to creating schedules in similar scenarios have avoided
this problem by using heuristics to find an approximately optimal solution 
in a shorter amount of time~\cite{burke_state_2004}.

We presented a formulation that includes both hard constraints to ensure the
schedule satisfies hospital and logistics requirements, and a multi-goal
objective function to satisfy soft constraints (work preferences of clinicians).
Although we restricted our application of the formulation to a set of
constraints for the particular needs of the case study (St.\ Michael's Hospital
Division of Infectious Diseases), our formulation can be adapted to various
clinical departments at different hospitals. The flexibility of our ILP allows
changing the number of services provided in a division, the length of a work
block, clinicians' preference for block to weekend adjacency as well as
clinicians' requests for time off.

When comparing the optimal schedule generated by our tool to the
manually-created schedules at St.\ Michael's Hospital, we found that the ILP
formulation was always able to find an optimal schedule satisfying all required
hard constraints, unlike the manual schedule, which often did not satisfy all
constraints. Moreover, due to the multi-goal objective function in
the ILP, the algorithm was able to fulfill the majority of clinician
preferences and requests, more so than the manually-created schedule. These
observations reinforce the benefits of automated tools when generating schedules
in hospital departments to balance the workload of clinicians and improve the
service provided to patients. The use of automated tools alleviates the time
spent on designing the schedule by hand, and provides clinical departments with
a more fair distribution of work that helps improve the overall satisfaction of
both employees and patients~\cite{silvestro_evaluation_2000}.

In our simulations, we also found that increasing the number of requests per
clinician did not affect the runtime of the algorithm, highlighting the
flexibility of the tool to incorporate clinician preferences. Further, we saw
that the algorithm can accommodate an increase in time-horizon up to four years
with little impact on runtime, suggesting the algorithm can be used generate
schedules far in advance. A key limitation we identified was the sensitivity
of the runtime to larger numbers of services offered by a single department. 
One solution to mitigate the runtime issues created by a
larger number of services would be removing the constraint that prevents
assignment of consecutive blocks, followed by manual readjustment from the
generated schedule.
Overall, our sensitivity analyses using simulated data provided reassurance that
the ILP formulation can be applied to schedule clinicians across real-world
variability between clinical departments.
Next steps include expanding the generalizability of the tool beyond smaller
clinical departments to larger departments within and outside of health-care --
especially those that provide multiple services in parallel for patients and
other clients. As well, additional work can be done to incorporate other
clinician preferences, such as the ability to request time-on slots or preference
for time of year.
	\section{Acknowledgements}\label{sec:acknowledgements}
	The authors are grateful to Julie Veitch (St.\ Michael's Hospital) for her contributions to testing and
designing the scheduling software. SM is supported by a Canadian Institutes of
Health Research New Investigator Award. JK is supported by a Natural Sciences and 
Engineering Research Council doctoral award. DL conducted part of this project as a
Keenan Research Summer Student, Li Ka Shing Knowledge Institute, St.\ Michael's
Hospital, University of Toronto.

	\section{Funding}\label{sec:funding}
	The work was jointly funded by the Ontario Ministry of Science and Innovation Early Researcher Award Number ER17-13-043; the Division of Infectious Diseases, St.\ Michael's Hospital, University of Toronto; and the University of Toronto Work Study Program.
	
	\bibliographystyle{unsrt}
	\bibliography{references}

\begin{thebibliography}{10}

\bibitem{aickelin_improved_2006}
Uwe Aickelin, Edmund~K. Burke, and Jingpeng Li.
\newblock Improved {Squeaky} {Wheel} {Optimisation} for {Driver} {Scheduling}.
\newblock In Thomas~Philip Runarsson, Hans-Georg Beyer, Edmund Burke, Juan~J.
  Merelo-Guervós, L.~Darrell Whitley, and Xin Yao, editors, {\em Parallel
  {Problem} {Solving} from {Nature} - {PPSN} {IX}}, Lecture {Notes} in
  {Computer} {Science}, pages 182--191. Springer Berlin Heidelberg, 2006.

\bibitem{goel_truck_2012}
Asvin Goel, Claudia Archetti, and Martin Savelsbergh.
\newblock Truck driver scheduling in {Australia}.
\newblock {\em Computers \& Operations Research}, 39(5):1122--1132, May 2012.

\bibitem{gunther_combined_2010}
Maik Günther and Volker Nissen.
\newblock Combined {Working} {Time} {Model} {Generation} and {Personnel}
  {Scheduling}.
\newblock In Wilhelm Dangelmaier, Alexander Blecken, Robin Delius, and Stefan
  Klöpfer, editors, {\em Advanced {Manufacturing} and {Sustainable}
  {Logistics}}, Lecture {Notes} in {Business} {Information} {Processing}, pages
  210--221. Springer Berlin Heidelberg, 2010.

\bibitem{al-yakoob_mixed-integer_2007}
Salem~M. Al-Yakoob and Hanif~D. Sherali.
\newblock Mixed-integer programming models for an employee scheduling problem
  with multiple shifts and work locations.
\newblock {\em Annals of Operations Research}, 155(1):119--142, November 2007.

\bibitem{al-yakoob_column_2008}
S.~M. Al-Yakoob and H.~D. Sherali.
\newblock A column generation approach for an employee scheduling problem with
  multiple shifts and work locations.
\newblock {\em Journal of the Operational Research Society}, 59(1):34--43,
  January 2008.

\bibitem{alfares_simulation_2007}
H.~K. Alfares.
\newblock A {Simulation} {Approach} for {Stochastic} {Employee} {Days}-{Off}
  {Scheduling}.
\newblock {\em International Journal of Modelling and Simulation}, 27(1):9--15,
  January 2007.

\bibitem{chapados_retail_2011}
Nicolas Chapados, Marc Joliveau, and Louis-Martin Rousseau.
\newblock Retail {Store} {Workforce} {Scheduling} by {Expected} {Operating}
  {Income} {Maximization}.
\newblock In Tobias Achterberg and J.~Christopher Beck, editors, {\em
  Integration of {AI} and {OR} {Techniques} in {Constraint} {Programming} for
  {Combinatorial} {Optimization} {Problems}}, Lecture {Notes} in {Computer}
  {Science}, pages 53--58. Springer Berlin Heidelberg, 2011.

\bibitem{nissen_automatic_2010}
Volker Nissen and Maik Günther.
\newblock Automatic {Generation} of {Optimised} {Working} {Time} {Models} in
  {Personnel} {Planning}.
\newblock In Marco Dorigo, Mauro Birattari, Gianni~A. Di~Caro, René Doursat,
  Andries~P. Engelbrecht, Dario Floreano, Luca~Maria Gambardella, Roderich
  Groß, Erol Şahin, Hiroki Sayama, and Thomas Stützle, editors, {\em Swarm
  {Intelligence}}, Lecture {Notes} in {Computer} {Science}, pages 384--391.
  Springer Berlin Heidelberg, 2010.

\bibitem{horn_scheduling_2007}
M.~E.~T. Horn, H.~Jiang, and P.~Kilby.
\newblock Scheduling patrol boats and crews for the {Royal} {Australian}
  {Navy}.
\newblock {\em Journal of the Operational Research Society}, 58(10):1284--1293,
  October 2007.

\bibitem{laguna_modeling_2005}
Manuel Laguna and Terry Wubbena.
\newblock Modeling and {Solving} a {Selection} and {Assignment} {Problem}.
\newblock In Bruce Golden, S.~Raghavan, and Edward Wasil, editors, {\em The
  {Next} {Wave} in {Computing}, {Optimization}, and {Decision} {Technologies}},
  Operations {Research}/{Computer} {Science} {Interfaces} {Series}, pages
  149--162. Springer US, 2005.

\bibitem{azaiez_0-1_2005}
M.N. Azaiez and S.S. Al~Sharif.
\newblock A 0-1 goal programming model for nurse scheduling.
\newblock {\em Computers \& Operations Research}, 32(3):491--507, March 2005.

\bibitem{trilling_nurse_2006}
Lorraine Trilling, Alain Guinet, and Dominiue Le~Magny.
\newblock Nurse scheduling using integer linear programming and constraint
  programming.
\newblock {\em IFAC Proceedings Volumes}, 39(3):671--676, 2006.

\bibitem{widyastiti_nurses_2016}
Maya Widyastiti, Amril Aman, and Toni Bakhtiar.
\newblock Nurses {Scheduling} by {Considering} the {Qualification} using
  {Integer} {Linear} {Programming}.
\newblock {\em TELKOMNIKA (Telecommunication Computing Electronics and
  Control)}, 14(3):933, September 2016.

\bibitem{el_adoly_new_2018}
Ahmed~Ali El~Adoly, Mohamed Gheith, and M.~Nashat~Fors.
\newblock A new formulation and solution for the nurse scheduling problem: {A}
  case study in {Egypt}.
\newblock {\em Alexandria Engineering Journal}, 57(4):2289--2298, December
  2018.

\bibitem{aickelin_exploiting_2000}
Uwe Aickelin and Kathryn~A. Dowsland.
\newblock Exploiting problem structure in a genetic algorithm approach to a
  nurse rostering problem.
\newblock {\em Journal of Scheduling}, 3(3):139--153, 2000.

\bibitem{jan_evolutionary_2000}
A.~Jan, M.~Yamamoto, and A.~Ohuchi.
\newblock Evolutionary algorithms for nurse scheduling problem.
\newblock In {\em Proceedings of the 2000 {Congress} on {Evolutionary}
  {Computation}. {CEC}00 ({Cat}. {No}.00TH8512)}, volume~1, pages 196--203
  vol.1, July 2000.

\bibitem{kawanaka_genetic_2001}
H.~Kawanaka, K.~Yamamoto, T.~Yoshikawa, T.~Shinogi, and S.~Tsuruoka.
\newblock Genetic algorithm with the constraints for nurse scheduling problem.
\newblock In {\em Proceedings of the 2001 {Congress} on {Evolutionary}
  {Computation} ({IEEE} {Cat}. {No}.01TH8546)}, volume~2, pages 1123--1130 vol.
  2, May 2001.

\bibitem{jaszkiewicz_metaheuristic_1997}
Andrzej Jaszkiewicz.
\newblock A metaheuristic approach to multiple objective nurse scheduling.
\newblock {\em Foundations of Computing and Decision Sciences}, 22(3):169--184,
  1997.

\bibitem{abdennadher_nurse_1999}
Slim Abdennadher and Hans Schlenker.
\newblock Nurse scheduling using constraint logic programming.
\newblock In {\em {AAAI}/{IAAI}}, pages 838--843, 1999.

\bibitem{li_hybrid_2003}
Haibing Li, Andrew Lim, and Brian Rodrigues.
\newblock A {Hybrid} {AI} {Approach} for {Nurse} {Rostering} {Problem}.
\newblock In {\em Proceedings of the 2003 {ACM} {Symposium} on {Applied}
  {Computing}}, {SAC} '03, pages 730--735, New York, NY, USA, 2003. ACM.

\bibitem{burke_state_2004}
Edmund~K. Burke, Patrick De~Causmaecker, Greet~Vanden Berghe, and Hendrik
  Van~Landeghem.
\newblock The {State} of the {Art} of {Nurse} {Rostering}.
\newblock {\em Journal of Scheduling}, 7(6):441--499, November 2004.

\bibitem{goos_complexity_1996}
Tim~B. Cooper and Jeffrey~H. Kingston.
\newblock The complexity of timetable construction problems.
\newblock In Gerhard Goos, Juris Hartmanis, Jan Leeuwen, Edmund Burke, and
  Peter Ross, editors, {\em Practice and {Theory} of {Automated}
  {Timetabling}}, volume 1153, pages 281--295. Springer Berlin Heidelberg,
  Berlin, Heidelberg, 1996.

\bibitem{hammer_boolean_1968}
P.~L. Hammer and S.~Rudeanu.
\newblock {\em Boolean {Methods} in {Operations} {Research} and {Related}
  {Areas}}.
\newblock {\"O}konometrie und {Unternehmensforschung} {Econometrics} and
  {Operations} {Research}. Springer-Verlag, Berlin Heidelberg, 1968.

\bibitem{stanimirovic_linear_2011}
Ivan~P. Stanimirovic, Milan~Lj Zlatanovic, and Marko~D. Petkovic.
\newblock On the linear weighted sum method for multi-objective optimization.
\newblock {\em Facta Acta Univ}, 26(4):49--63, 2011.

\bibitem{mitchell_branch-and-cut_2002}
John~E. Mitchell.
\newblock Branch-and-cut algorithms for combinatorial optimization problems.
\newblock {\em Handbook of applied optimization}, 1:65--77, 2002.

\bibitem{shamir_efficiency_1987}
Ron Shamir.
\newblock The {Efficiency} of the {Simplex} {Method}: {A} {Survey}.
\newblock {\em Management Science}, 33(3):301--334, 1987.

\bibitem{landsman_scheduling}
David Landsman.
\newblock On-call scheduling tool for clinicians.
\newblock \url{https://github.com/c-uhs/scheduler}, 2019.

\bibitem{johnjforrest_coin-or/cbc:_2019}
{johnjforrest}, Stefan Vigerske, Ted Ralphs, Haroldo~G. Santos, Lou Hafer,
  Bjarni Kristjansson, {jpfasano}, Edwin Straver, Miles Lubin, {rlougee},
  {jpgoncal1}, {h-i-gassmann}, and Matthew Saltzman.
\newblock coin-or/{Cbc}: {Version} 2.9.9, 2019.

\bibitem{silvestro_evaluation_2000}
Rhian Silvestro and Claudio Silvestro.
\newblock An evaluation of nurse rostering practices in the {National} {Health}
  {Service}.
\newblock {\em Journal of Advanced Nursing}, 32(3):525--535, 2000.

\end{thebibliography}
\end{document}